\documentclass[11pt,a4paper]{article}
\usepackage[hyperref]{acl2017}
\usepackage{times}
\usepackage{latexsym}

\usepackage{amssymb}
\usepackage{amsmath}
\usepackage{centernot}
\usepackage{mathtools}
\usepackage{graphicx}
\usepackage{verbatim}
\usepackage{mathabx}

\usepackage{url}

\aclfinalcopy 

\title{Fixing the Infix: Unsupervised Discovery of Root-and-Pattern Morphology}

\author{Tarek Sakakini\\
  University of Illinois
   \\\And
  Suma Bhat\\
  University of Illinois
   \\\And
   Pramod Viswanath\\
  University of Illinois
   }

\date{}

\begin{document}
\maketitle

\begin{abstract}
We present an unsupervised and language-agnostic method for  learning root-and-pattern morphology in Semitic languages. 
 We harness the syntactico-semantic information in distributed word representations to solve the long standing problem of root-and-pattern discovery in Semitic languages. The root-and-pattern morphological rules we learn in an unsupervised manner are validated by native speakers in Arabic, Hebrew, and Amharic.  Further, the rules are used to construct an unsupervised root extractor called JZR, which we show to compare favorably with ISRI, a widely used and carefully engineered root extractor for Arabic. 
\end{abstract}

\section{Introduction}


Morphological analysis is a core task  in natural language processing and hence of central interest in computational linguistics \cite{Kurimo:2010:MCC:1870478.1870489} with widespread applications benefits  \cite{korenius2004stemming,skorkovska2012application,seddah2010lemmatization,semmar2006using,larkey2002improving}. Computational approaches to morphological analysis for lemmatization and stemming rely on knowledge of the mapping from inflected forms to lemmas. However, the more widely used systems that learn these mappings are restricted to handle only \textit{concatenative} morphology, learning overt affixes and their interactions with stems  \cite{creutz2005unsupervised,soricut2015unsupervised,lee2011modeling,goldsmith2000linguistica,pasha2014madamira,buckwalter2004buckwalter}.  As an example, consider the agentive formation in Arabic (Table \ref{tab: examples}): (to write, ktb) in the root form to (writer, kAtib) in the agent form. Unlike in English, where the agentive formation occurs  as a concatenation of the suffix ``er'',  in Arabic, the process involves applying a specific pattern (C$_1$AC$_2$iC$_3$, in the notation of Section \ref{sec: explain}) to the root ktb to form the word kAtib. Such processes, which constitute a significant component in Semitic languages, are not captured by existing unsupervised systems. The focus of this study is a computational approach to an important non-concatenative process--that of root-and-pattern morphology. 


\begin{table}
\centering
\begin{tabular}{|c|c|c|}
\hline
Root & Derived & Template\\\hline
{\em ktb} & kAtib (writer) & C$_1$AC$_2$iC$_3$\\\hline
{\em ktb} & maktab (desk) & maC$_1$C$_2$aC$_3$\\\hline
\end{tabular}
\caption{Illustration of Arabic root-and-pattern rules on the root {\em ktb} (to write).}
\label{tab: examples}
\end{table}

Current approaches for root-and-pattern morphology involve  rule-based root extraction \cite{khoja1999stemming,taghva2005arabic,ababneh2012building,el2011accuracy,alhanini2011enhancement,Al-Shalabi:1998:CMS:1621753.1621765}, and others that require training data of word-root pairs \cite{attia2016power,4115547}, all developed for processing Arabic. The resource-intensive nature of these methods  not only limits their wider applicability to other Semitic languages but also prevents from handling the productive process more generally.

This paper describes a language-agnostic (within the Semitic language family)  algorithm that learns root-and-pattern, as well as concatenative morphology in an unsupervised manner resulting in a root extractor. A novel aspect of our approach is the use of word semantics made possible by relying on distributed word representations. Such a reliance on word representations permits a different mechanism of morpheme segmentation, which is being performed by an abstraction that is not obvious at the surface level.  Qualitative analyses show perfect rules acquired for the 3 languages.  Quantitative analyses of root extraction on Arabic show performance comparable to carefully engineered systems.

\section{Root-and-Pattern Morphology Illustration}
\label{sec: explain}
We explain root-and-pattern morphology with an example. Our examples will take the form (English translation, English transliteration, Consonant-only transliteration), and transliteration will follow the standard in \cite{lagally1992arabtex}. In Semitic languages, a pattern (bound morpheme) is applied to a root (free morpheme) to generate a  word. In general, the roots consist of 3 consonants; example: (to write, kataba, ktb) \cite{Aljlayl:2002:ASI:584792.584848}. The patterns encode the placement of the 3 consonants with respect to the added letters, an example of which  is, C$_1$AC$_2$iC$_3$. Here, the vowel ``A'' is placed between the first and the second consonant and the vowel ``i''  between the second and the third consonant. Applying this pattern to the 3-consonant root in our example, the resulting word is (writer, kAtib, ktb). This pattern encodes the semantic role of ``Agent''  and hence kAtib is the agent of the verb kataba.


\section{Learning of Morphological Rules}

The task of morphological rule learning is, given a corpus, we would like to extract patterns that govern root-and-pattern morphology as well as affixes that govern concatenative morphology. Our proposed method assumes three models. An orthographic model for concatenative morphology and an orthographic model  for root-and-pattern morphology signal all possible morphological (candidate) rules and all possible (candidate) pairs of morphologically related words from an orthographic perspective. Then a semantic model validates whether a candidate rule is morphologically valid and whether a pair of words is morphologically related.

Existing approaches towards root-and-pattern morphology in the literature tend to be a set of orthographic rules defining the model; the heavy reliance on expert linguistic rules naturally leads to poor coverage. Besides, the orthographic rules disregard the semantically motivated morphological transformation. For example, in English, the pair of words (on, only) shows orthographically a morphological relation via the suffix ``ly''. Disregarding semantic knowledge in such a situation would lead to a false morphological analysis. This motivates our approach which models the relation between roots, patterns, and words through a combination of orthographic rules and semantic information available in distributed word representations.

\subsection{Orthographic Models}
\label{sec: typo model}

To detect candidate morphological rules from an orthographic standpoint, we define two models, one for concatenative morphology, and one for root-and-pattern morphology.

\subsubsection{Concatenative Morphology}

In concatenative morphology, affixes attach to a stem \cite{siegel1974topics}. Thus we model concatenative morphology as the process of deletions from one end of the word followed by insertions on the same end.  For example, the words (desk, maktab, mktb) and (the desk, almaktab, lmktb) are considered morphologically related from an orthographic standpoint whereby there was 0 deletions in the beginning of the word followed by two insertions (a,l). This would contribute to the candidate rule (prefix, $\phi$, al). Another example would be the pair of words (the desk, almaktab, lmktb) and (and desk, wamaktab, wmktb). Based on this model, they would be linked via two deletions (a,l) and two insertions (w,a), and this pair would contribute to the rule (prefix, al, wa). We found a maximum of 6 insertions and deletions to be sufficient to model the concatenative relation. 

\subsubsection{Root-and-Pattern Morphology}

All Semitic languages share the following two properties of relations between roots and their pattern-derived words on an orthographic level \cite{greenberg1950patterning}:
\begin{itemize}
\item Roots are predominantly tri-literal.
\item The root consonants (C$_1$, C$_2$, C$_3$) retain their relative order after combining with a pattern to form a word.
\end{itemize}
For example, in Table \ref{tab: examples}, the word kAtib would be compared to the tri-literal root ktb, and consequently, the template C$_1$AC$_2$iC$_3$ would be considered as a candidate root-and-pattern template.

\subsection{Semantic Model}
\label{sec: sem model}
Not every pair of words obeying the orthographic properties, constitutes a morphologically related pair. As an example, (gone, rA.h) and (wounds, jirA.h) are connected orthographically via the rule (prefix, $\phi$, ji) although they're not morphologically related. In order to ascertain that a pair of words obeying the orthographic properties are indeed morphologically related, we make use of the semantic relationship between the words via word representations. This is because word representations and their geometry in the vector space have been shown to encode syntactico-semantic relations \cite{mikolov2013linguistic}. 
For example, consider the two pairs: (1) ((wrote, kataba, ktb), (desk, maktab, mktb)), and,
(2) ((played, la`iba, l`b),  (playing field, mal`ab, ml`b)). We can verify that $\vec{v}_{kataba} - \vec{v}_{maktab} \approx \vec{v}_{la`iba} - \vec{v}_{mal`ab}$, to confirm that orthographically related word pairs are indeed morphologically related, here through the pattern (maC$_1$C$_2$aC$_3$) which means the place of the action.  On the other hand, alluding to the previous example $r$ = (prefix, $\phi$, ji), difference vectors of elements of the support set of this rule would not show such regularities in the vector space.


\subsection{Algorithmic Implementation}

 Our approach is inspired by \cite{soricut2015unsupervised}, which, being limited to concatenative morphology,  serves as the primary mechanism to generate the concatenative morphological rules for the languages in our study. Our contribution is in adapting their approach to handle non-concatanative morphology. Our system  is implemented in 4 steps:
 

\begin{enumerate}
\item Generation of the vocabulary set $V$ and word embeddings from corpus (\textbf{Word representation});
\item Generation of candidate (root, derived) pairs grouped into sets based on the underlying orthographic transformation. These sets represent the morphological rules in our system. (\textbf{Candidate generation});
\item Validation of candidate morphological rules using the semantic and orthographic model (\textbf{Validation of candidate rules}).
\item  Validation of candidate rule elements as morphologically related pairs of words  (\textbf{Validation of rule elements}).
\end{enumerate}
The result is a set of morphological rules and their elements, scored semantically and orthographically.

\textbf{Step 1: Word representation}. We rely on  a large enough corpus to generate a vocabulary set $V$ made up of all the word types appearing in the corpus, which is then used to generate word embeddings for the vocabulary using an  algorithm such as Word2Vec \cite{mikolov2013efficient} or Glove \cite{pennington2014glove}.

\textbf{Step 2: Candidate generation}. All pairs of words morphologically related by the same orthographic transformation are grouped into one set to represent a candidate morphological rule. For example, the concatenative rule $r$ = (prefix, $\phi$, ``al'') would be represented by the support set SS$_r$ = \{(maktab, almaktab), (jaras, aljaras), ...\}. Similarly, the rule $r$ = (maC$_1$C$_2$aC$_3$), shown in Table \ref{tab: examples}, would have SS$_r$ = \{(ktb, maktab), (l'b, mal'ab), ...\}.  We should note that not all generated rules in this step are valid morphologically. We refer to the set of all rules as $R$.

\textbf{Step 3: Validation of candidate rules}. Due to the ``overgeneration" of candidate rules in the previous step, we prune the  rules using a semantic and an orthographic score. Formally,  for a given rule $r$ with support set SS$_r$, score$_{\text{r\_sem}}$($r$) = $|$\{($w_1$, $w_2$), ($w_3$, $w_4$) $\in$ SS$_r$ $|$ cos($w_4$, $w_2$ - $w_1$ + $w_3$) $>$ t$_{\text{cos}\_\text{sim}}$\}$|$ divided by $|\text{SS}_r|^2$, where t$_{\text{cos}\_\text{sim}}$ is a threshold appropriately chosen. Also, for every rule $r$, $|$SS$_r$$|$ reflects the quality of $r$ from an orthographic point of view. This is captured via score$_{\text{r\_orth}}$($r$) = $|$SS$_r$$|$. 


\textbf{Step 4: Validation of rule elements}. The orthographic model not only overgenerates candidate rules but also overgenerates pairs of words belonging to a specific rule. As an example, consider the words (to go, zhb) and (religion, mazhab) --  this pair belongs to the valid root-and-pattern rule guided by the pattern maC$_1$C$_2$aC$_3$, and yet, the words are not morphologically related. To score instances within a rule $r$ as valid instances we deploy this semantic score: score$_{\text{w\_sem}}$(($w_1$, $w_2$) $\in$ $r$) = $|$\{($w_3$, $w_4$) $\in$ SS$_r$ $|$ cos($w_4$, $w_2$ - $w_1$ + $w_3$) $>$ t$_{\text{cos}\_\text{sim}}$\}$|$ divided by $|$SS$_r$$|$. In other words, we check how well the difference vector of the pair of interest fits with the difference vectors of other pairs within the rule support set SS$_r$. 






\section{JZR: Root Extractor}
\label{sec: root extractor}



We cast the root extraction task as an iterative optimization problem. Let R$_{\text{add}}$ be all concatenative rules of the form (affix, $\phi$, affix added), as well as all root-and-pattern rules, and R$_{\text{rep}}$ be all concatenative rules of the form (affix, affix deleted, affix added). Given a word $w$ whose root needs to be extracted, we search for rule $r^{*}$, the solution to the following optimization problem:


\begin{equation*}
\begin{aligned}
& \underset{r}{\text{max}}
& & \text{score}_{\text{w\_sem}}(r) \\
& \text{subject to}
& & r \in \text{R}_\text{add} \\
&&& \text{score}_{\text{r\_sem}}(r) > t_{\text{r}\_\text{sem}}\\
&&& \text{score}_{\text{r\_orth}}(r) > t_{\text{r}\_\text{orth}}\\
&&& \text{score}_{\text{w\_sem}}((w', w) \in \text{SS}_r) > \text{t}_\text{w\_sem}\\
\end{aligned}
\end{equation*}

In the above optimization problem, t$_{\text{r}\_\text{sem}}$, t$_{\text{r}\_\text{orth}}$, t$_\text{w\_sem}$ are tunable hyperparameters. The solution $r^*$ and $w$ uniquely identify $w'$. Thus the system extracts $w'$ and iterates over $w'$ until it reaches a triliteral word. At any stage, if the optimization problem is infeasible we repeat it over R$_\text{rep}$ instead of R$_\text{add}$. Note that we only consider rules which result in a $w'$ of length less than $w$ to correctly model the chain of morphological processes as well to guarantee convergence of algorithm.

The system presented here is readily adaptable to other morphological tasks, such as morpheme segmentation and morphological reinflection and not discussed in this paper. 

\section{Experiments and Results}

In this section, we evaluate our method in multiple ways on 3 different languages (Arabic, Hebrew, Amharic). We first evaluate our method qualitatively by checking the root-and-pattern rules it discovers in the 3 languages. Then we evaluate JZR  quantitatively by comparing its accuracy against the widely used ISRI Arabic root extractor \cite{taghva2005arabic} on a sample of 1200 words.

\subsection{Experimental Setup}

For each of the three languages, we use the readily available Polyglot\footnote{https://sites.google.com/site/rmyeid/projects/polyglot} word representations and its vocabulary. These 64d word representations were created based on the Wikipedia in the respective language. Arabic and Hebrew embeddings were limited to the top 100K words whereas Amharic, being a low-resource language was restricted to the top 10K words. Our hyperparameters (tuned to ISRI output) were set to the following: t$_{\text{r}\_\text{sur}}$ = 20, t$_{\text{r}\_\text{sem}}$ = 0.1 , t$_{\text{w\_sem}}$ = 0.1, t$_{\text{cos}\_\text{sim}}$  = 0.5. 

\subsection{Evaluation of Root-and-Pattern rules}

We validate root-and-pattern rules across the 3 languages by ranking the rules in terms of $\text{score}_{\text{r\_sem}}(r)$ and have a native speaker of each language evaluate the top 30 rules.  All 30 rules in each of the 3 languages were deemed correct, validating the language-agnosticity and performance of our unsupervised approach.


\subsection{Intrinsic Evaluation on Arabic}

In this experiment, we consider 3 Arabic root extractors: JZR, JZR (limited), and ISRI, and evaluate them on a sample of 1200 words. JZR (limited) is a version of JZR limited to concatenative morphology. Comparison against it reflects the added value of the discovered root-and-pattern rules.

From the results (summarized in Table \ref{tab: root extra results}), we notice: (1) JZR compares favorably with ISRI, a carefully engineered rule-based and language-specific root extractor (2) Limiting JZR to concatenative morphology led to an 11.5\% relative drop in scores. This reflects the significance of the non-concatenative rules captured by JZR. We also claim that this drop is an underestimate of the significance of the learned root-and-pattern rules. We discuss this claim in detail  in Appendix A, along with further analyses of the results and sample outputs.

\begin{table}
\centering
\begin{tabular}{|c|c|c|}
\hline
JZR & JZR (limited) & ISRI\\\hline
51.88\% & 45.63 \% & 61.63\% \\\hline
\end{tabular}
\caption{Performance of the three root extractors on a sample of 1200 Arabic words.}
\label{tab: root extra results}
\end{table}


\section{Conclusion}

This work  presents an  unsupervised method for the discovery of root-and-pattern morphology in Semitic languages. The discovered rules are used to extract Semitic roots, which are the basic units of these languages. Intrinsic and extrinsic evaluations of these rules allow us to validate our pattern discovery method as well as our root extraction method (JZR), with performance not too far from a rule-based language-specific (in this case Arabic) root extractor. 

\pagebreak

\bibliography{acl2017}
\bibliographystyle{acl_natbib}

\appendix

\section{Discussion}
\label{sec: discussion}

We believe that the drop in performance when limiting JZR to concatenative morphology underestimates the abundance of root-and-pattern morphology in Arabic. The reason is that the way we define concatenative morphology could capture root-and-pattern morphology under strict conditions. For the purpose of illustration we have collected 5 representative examples into Table \ref{tab: root ext ex}, where correct roots are boldfaced. To illustrate this underestimate, consider as an example, the second word in Table \ref{tab: root ext ex}. All extractors were able to get the second example correct. Although, the word was derived using a pattern, JZR (limited) was still able to get it right since the stem change was close to the edge of the word. To illustrate this further, a word like kitAb is stripped first of the ``i'' using the rule (pre, ki, k), and similarly stripped of the ``A'' using a suffix rule. These cases are limited, since to extract such a rule, this pattern should appear frequently with a word starting (ending) with a ``k'' (``b''). The first example  shows how the limitation to concatenative morphology prevented JZR (limited) from removing the pattern, leading to a non-root word. In the third example, JZR fails for using a valid rule on an invalid pair of words, which reflects the imperfections in the word embeddings' linear structure.

For purposes of comparison, we also show one-to-one comparisons of performance in Table  \ref{tab: one-to-one}. Two key takeaways arise in this table. First, JZR (limited) never performs better than JZR, which shows the precision of discovered root-and-pattern rules. Second,  JZR performed better than JZR (limited) on 63 occasions due to the discovery of root-and-pattern morphology. Moreover, it is interesting to see that on multiple occasions JZR performed better than ISRI, which shows that rule-based methods are insufficient and unsupervised methods are needed to fill the gap.

\begin{table}
\centering
\begin{tabular}{|c|c|c|c|}
\hline
Word & JZR & JZR (limited) & ISRI\\\hline
lilta`Ayu\^s & {\bf `A\^sa} & ta`Ayu\^s & `ay\^s\\\hline
fasAdaN & {\bf fasada} & {\bf fasada}& {\bf fasada}\\\hline
lilkusUr & sUr & sUr& {\bf kasara}\\\hline
.hukkAmaN & {\bf .hakama} & {\bf .hakama}& a.hkAm\\\hline
wayamta.s.s & yamut & yamut& mta.s\\\hline
\end{tabular}
\caption{Comparison of roots extracted using JZR against NLTK's ISRI stemmer. Correct outputs are boldfaced.}
\label{tab: root ext ex}
\end{table}

\begin{table}
\centering
\begin{tabular}{|c|c|c|c|}
\hline
     & JZR & JZR (limited) & ISRI\\\hline
JZR & 0 & 63 & 126\\\hline
JZR (limited) & 0 & 0 & 120\\\hline
ISRI & 224 & 281 & 0\\\hline
\end{tabular}
\caption{One-to-one comparison of extractors. The number in the cell shows how many times the extractor in that row performed better than the extractor in the column.}
\label{tab: one-to-one}
\end{table}

\end{document}